\documentclass{article}
\usepackage{ijcai19}

\usepackage{times}
\usepackage{soul}
\usepackage{url}
\usepackage[hidelinks]{hyperref}
\usepackage[utf8]{inputenc}
\usepackage[small]{caption}
\usepackage{graphicx}
\usepackage{amsmath}
\usepackage{booktabs}
\usepackage{algorithm}
\usepackage{algorithmic}
\usepackage{amsfonts}
\title{Deep Multi-Agent Reinforcement Learning with Discrete-Continuous Hybrid Action Spaces}

\author{
	Haotian Fu$^1$
	\and
	Hongyao Tang$^1$\and
	Jianye Hao$^1$\and
	Zihan Lei$^2$\and
	Yingfeng Chen$^2$\and
	Changjie Fan$^2$\\
	\affiliations
	$^1$Tianjin University\\
	$^2$Fuxi AI Lab in Netease
	\emails
	\{haotianfu, bluecontra, jianye.hao\}@tju.edu.cn,\\
	\{leizihan, chenyingfeng1, fanchangjie\}@corp.netease.com
}

\begin{document}
	
	\maketitle
	
	\begin{abstract}
		
		Deep Reinforcement Learning (DRL) has been applied to address a variety of cooperative multi-agent problems with either discrete action spaces or continuous action spaces. However, to the best of our knowledge, no previous work has ever succeeded in applying DRL to multi-agent problems with discrete-continuous hybrid (or parameterized) action spaces which is very common in practice. Our work fills this gap by proposing two novel algorithms: Deep Multi-Agent Parameterized Q-Networks (Deep MAPQN) and Deep Multi-Agent Hierarchical Hybrid Q-Networks (Deep MAHHQN). We follow the centralized training but decentralized execution paradigm: different levels of communication between different agents are used to facilitate the training process, while each agent executes its policy independently based on local observations during execution. Our empirical results on several challenging tasks (simulated RoboCup Soccer and game Ghost Story) show that both Deep MAPQN and Deep MAHHQN are effective and significantly outperform existing independent deep parameterized Q-learning method. 
		
	\end{abstract}
	
	\section{Introduction}
	
	Reinforcement learning (RL) has recently shown a great success on a variety of cooperative multi-agent problems, such as multiplayer games~\cite{DBLP:journals/corr/PengYWYTLW17}, autonomous cars~\cite{Cao2013AnOO} and network packet delivery~\cite{DBLP:journals/sensors/YeZY15}.
	
	In many such settings, it is necessary for agents to learn decentralized policies due to the partial observability and limited communication. Fortunately, we can learn such policies using the paradigm of centralized training and decentralized execution. Forester et al.~\shortcite{DBLP:conf/nips/FoersterAFW16} developed a decentralized multi-agent policy gradient algorithm; Lowe et al.~\shortcite{Lowe2017MultiAgentAF} extended DDPG~\cite{Lillicrap2016ContinuousCW} to multi-agent setting with a centralized Q-function; Rashid et al.~\shortcite{Rashid2018QMIXMV} employs a Qmix network that estimates joint action-values as a complex non-linear combination of per-agent action values that condition on local observations only.
	
	These popular multi-agent reinforcement learning methods all require the action space to be either discrete or continuous. However, very often the action space in real world is discrete-continuous hybrid, such as Robot soccer~\cite{Hausknecht2016HalfFO,DBLP:conf/aaai/MassonRK16} and Real Time Strategic (RTS) games~\cite{Xiong2018ParametrizedDQ}. In such settings, each agent usually needs to choose a discrete action and the continuous parameters associated with it at each time step. An obvious approach to handling this is to simply approximate hybrid action spaces by a discrete set or relax it into a continuous set~\cite{Hausknecht2016DeepRL}. However, such approaches suffer from a number of limitations: for the continuous part of hybrid action, establishing a good approximation usually requires a huge number of discrete actions; for the discrete part of hybrid action, relaxing them into a continuous set might significantly increase the complexity of the action space.
	
	An alternative and better solution is to learn directly over hybrid action spaces. Following this direction,  Xiong et al.~\shortcite{Xiong2018ParametrizedDQ} proposed Parameterized Deep Q-Network (P-DQN) for single-agent learning in hybrid action spaces without approximation or relaxation by seamlessly integrating DQN~\cite{Mnih2013PlayingAW} and DDPG~\cite{Lillicrap2016ContinuousCW}. However P-DQN cannot be directly applied to multi-agent settings due to the non-stationary property in multi-agent environments. In multi-agent settings, explicit coordination mechanism among agents' hybrid action spaces needs to be introduced.
	
	In this work, we propose two approaches to address cooperative multi-agent problems in discrete-continuous hybrid action spaces based on centralized training and decentralized execution framework. The first approach, Deep Multi-Agent Parameterized Q-networks (Deep MAPQN), extends the architecture of P-DQN to multi-agent settings by leveraging the idea of Qmix~\shortcite{Rashid2018QMIXMV}. Our algorithm utilizes a joint action-value function to update policies of hybrid actions for all agents. However, Deep MAPQN requires to compute continuous parameters for all the discrete actions of all agents and thus may suffer from high computational complexity when the discrete part of hybrid action space has large dimensions. To alleviate this problem, we propose the second approach, Deep Multi-Agent Hierarchical Hybrid Q-networks (Deep MAHHQN). In contrast to Deep MAPQN, Deep MAHHQN only needs to calculate continuous parameters of the optimal discrete action for each agent. In addition, Deep MAHHQN further alleviates the non-stationary issue of multi-agent environments by realizing centralized training fashion at both action levels and augmenting each centralized training framework with information about policies of other action levels. Empirical results on standard benchmark game Half Field Offense (HFO) and a large-scale online video game \textit{Ghost Story} show the superior performance of our approaches compared to independent P-DQN.
	\section{Background}
	\subsection{Cooperative Stochastic Game}
	In this work, we consider a $Cooperative$ $Stochastic$ $Game$~\cite{DBLP:conf/aaaiss/WeiWFL18} in partially observable settings modeled as a tuple $\left \{S,U,r,P,\gamma , H,N\right \}$ . This game for $N$ agents is defined by a set $\mathcal{S}$ of states describing the possible configurations of all agents, a set of observations $\mathcal{O}_{1},..., \mathcal{O}_{N}$ for each agent and a joint action space of N agents defined as $U = \mathcal{A}_{1} \times \cdots \times \mathcal{A}_{N}$. At each time step, each agent $i$ chooses an action $a_{i} \in \mathcal{A}_{i}$ using policy $\pi_{i}$ and constitutes a joint action $\vec{a}  \in  U$, which produces the next state $s'$ following the state transition function $P(s'|s, \vec{a})$: $S\times U \times S \rightarrow [0,1]$. All agents share the same reward function $r(s,\vec{a})$: $S\times U \rightarrow \mathbb{R}$. $\gamma$ is the discount factor and $H$ is the time horizon.  
	\subsection{Deep Multi-agent Reinforcement Learning}
	\label{sec:ma}
	Multi-agent learning has been investigated comprehensively in both discrete action domains and continuous action domains under framework of centralized training and decentralized execution. MADDPG~\cite{Lowe2017MultiAgentAF} and Qmix~\cite{Rashid2018QMIXMV} are representative ones for discrete and continuous action domains..
	
	MADDPG mainly focuses on multi-agent problems with continuous action spaces. The core idea is to learn a centralized critic $Q_{I}^{\mu}(x,a_{1},\cdots,a_{N})$ for each agent which conditions on global information. $Q_{I}^{\mu}(x,a_{1},\cdots,a_{N})$ takes as input the actions of all agents, $a_{1},\cdots,a_{N}$, in addition to global state information $x$ (i.e., $x=(o_{1},\cdots,o_{N})$) and outputs the centralized action-value for agent $i$.
	
	Qmix employs a mixing network that estimates joint action-values $Q_{tot}$ as a complex non-linear combination of per-agent action value that conditions only on local observations. The weights of the mixing network are produced by separate hypernetworks which take the global state information as input. Importantly, Qmix ensures a global $argmax$ performed on $Q_{tot}$ yields the same result as a set of individual $argmax$ operations performed on each agent's action-value $Q_{i}$ by a monotonicity constraint: 
	\begin{equation}
	\frac{\partial Q_{tot}}{\partial Q_{i}}\geq  0,
	\end{equation}
	Unfortunately, such multi-agent algorithms can only be applied to either discrete action space or continuous action space. For multi-agent problems with hybrid action space, explicit coordination mechanism among agents needs to be introduced.
	\subsection{Parameterized Deep Q-Networks}
	To handle reinforcement learning problems with hybrid action space, Xiong et al.~\shortcite{Xiong2018ParametrizedDQ} propose a new framework called Parameterized Deep Q-Networks (P-DQN). The core idea is to update the discrete-action and continuous-action policies separately combining the structure of DQN~\cite{Mnih2013PlayingAW} and DDPG~\cite{Lillicrap2016ContinuousCW}. P-DQN first chooses low-level parameters associated with each high level discrete action, then figures out which discrete-continuous hybrid action pair maximizes the action-value function.
	
	More concretely, we can define the discrete-continuous hybrid action space $\mathcal{A}$ as:
	\begin{equation}
	\mathcal{A}=\left \{ (k,x_{k})|x_{k}\in \mathcal{X}_{k} \textrm{ for all }k\in [K] \right \},
	\end{equation}
	where $[K] = \left \{ 1,\cdots ,K \right \}$ is the discrete action set, $\mathcal{X}_{k}$ is a continuous set for all $k \in [K]$. Then we can define a deterministic function which maps the state and each discrete action $k$ to its corresponding continuous parameter $x_{k}$:
	\begin{equation}
	x_{k}=\mu_{k}(s;\theta),
	\end{equation} 
	where $\theta$ are weights of the deterministic policy network. We further define an action-value function $Q \left ( s, k, x_{k}; \omega \right )$ which maps the state and hybrid actions to real values. Here $\omega$ are weights of the value network.
	
	P-DQN updates the action-value function $Q$ by minimizing the following loss:
	\begin{equation}
	l^{Q}(\omega)=\frac{1}{2}\left [ Q(s,k,x_{k}; \omega)-y \right ]^{2},
	\end{equation}
	where $y=r+\max_{k \in [K]} \gamma Q(s',k,\mu_{k}(s';\theta); \omega),$ and $s'$ denotes the next state after taking hybrid action $(k, x_{k})$. The policy $\mu_{k}$ for the continuous part is updated by minimizing the following loss with parameters $\omega$ fixed:
	\begin{equation}
	l^{\Theta }(\theta)=-\sum_{k=1}^{K}Q(s,k,\mu_{k}(s;\theta); \omega)
	\end{equation} 
	Note that here the action value function $Q\left ( s, k, x_{k}; \omega \right)$ mainly plays two roles. First, it outputs the greedy policy for discrete action (which is consistent with DQN) . Secondly, it works as the critic value for different associated continuous parameters like DDPG which provides gradients for updating policies in continuous action space.

	\section{Methods}
	To handle hybrid action spaces in multi-agent settings, one natural approach is to adopt the independent learning paradigm and equip each agent with an independent P-DQN algorithm. However, one major issue is that each agent's policy changes during training, resulting in the non-stationarity of environments. As we will show in our experiments in Section 4, independent P-DQN does not perform well in practice.
	
	In this paper, we propose two novel deep multi-agent learning methods for hybrid action spaces, Deep MAPQN and Deep MAHHQN. By leveraging the current state-of-the-art single-agent deep RL for hybrid action spaces and coordination techniques for multi-agent learning, both Deep MAPQN and Deep MAHHQN can support multiple agents to learn effective coordination policies directly over the hybrid action spaces. 
	
	\subsection{Deep Multi-Agent Parameterized Q-Networks (Deep MAPQN)}
\begin{figure}[t]
		\centering
		\includegraphics[width=0.99\linewidth]{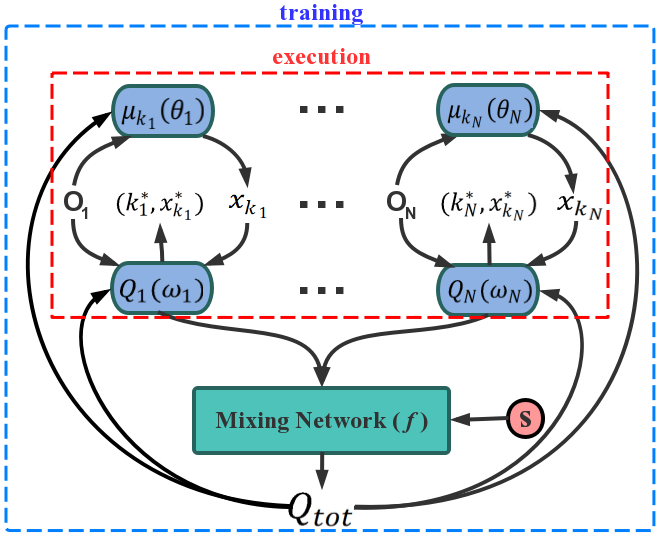}
		\caption{The overall Deep MAPQN structure}
		\label{fig:100}
	\end{figure}

	The first algorithm is a natural extention of single-agent P-DQN. We leverage Qmix~\cite{Rashid2018QMIXMV} architecture to coordinate the policy update over hybrid action spaces among agents. The overall structure of Deep MAPQN is shown in Figure \ref{fig:100}.
	
	For each agent, we adopt the same settings in P-DQN. Concretely, considering a game with $N$ agents, each agent $i$ uses a deterministic policy network $\mu_{k_{i}}(\theta_{i})$ and an action value network $Q_{i}(\omega_{i})$ to output hybrid action $(k_{i}^{*}, x_{k_{i}}^{*})$. The deterministic policy network $\mu_{k_{i}}(\theta_{i})$ takes each agent's observation $o_{i}$ as input and outputs the optimal continuous parameters $x_{k_{i}}$ for all the discrete actions $k_{i}\in \left \{ 1,\cdots ,K_{i} \right \} $. Then the action value network $Q_{i}$ outputs the optimal hybrid action by:
	\begin{equation}
	(k_{i}^{*}, x_{k_{i}}^{*}) = argmax_{(k_{i},x_{k_{i}})}Q_{i}\left ( s_{i}, (k_{i}, x_{k_{i}}); \omega _{i} \right ),
	\end{equation} 
	where $\omega_{i}$ are parameters for action value network of agent $i$.
	
	To achieve coordinated update among agents' action value networks, we utilize a mixing network and produce a fully centralized state-action value function $Q_{tot}$ which can facilitate the coordinated update of the decentralized policies in hybrid action spaces. The mixing network consists of a feedforward neural network and separate hypernetworks. The hypernetworks take the global state $\textbf{s}$ as the input and output the weights of the feedforward network. The feedforward network takes each agent's output $Q_{i}$ as input and mixing them monotonically, producing the joint action value denoted by $Q_{tot}$.\footnote{The detailed structure for the mixing network can be found at: \url{https://bit.ly/2Eaci2X}.} Here we define the mixing network as a non-linear complex function $f$ and denote this process by:
	\begin{equation}
	Q_{tot} = f(\textbf{s},Q_{1},\cdots,Q_{N};\omega_{mix})
	\end{equation}
	We update the mixing network weights $\omega_{mix}$ along with each agent's action value network weights $\omega_{i}$ by minimizing: 
	\begin{equation}
	\mathcal{L}(\omega ) = \mathbb{E}_{s,\vec{k},\vec{x}_{k},r,s'\sim \mathcal{D}}[y^{tot} - Q_{tot}( \textbf{s}, \vec{k},\vec{x}_{k})]^{2},
	\end{equation}
	where $y^{tot} = r + \gamma \max_{\vec{k}',\vec{x}_{k'}}Q_{tot} (\textbf{s}',\vec{k}', \vec{x}_{k'}(\theta')),$ and $(\vec{k}, \vec{x}_{k})$ is the joint action, $\theta'$ are parameters of target policy networks. Our framework for computing $Q_{tot}$ ensures off-policy learning updates while each agent can still choose the greedy action with respect to its own $Q_{i}$ in a decentralized fashion. This is because a global argmax on $Q_{tot}$ is equivalent with argmax on each $Q_{i}$ as explained in Section \ref{sec:ma}.
	
	Finally we need to compute gradients for deterministic policy network. We first take the sum of Q-values of all the discrete actions for each agent $i$:
	\begin{equation}
	\widehat{Q}_{i}=\sum_{k_{i}=1}^{K_{i}}Q_{i}\left ( s_{i}, k_{i}, x_{k_{i}}; \omega _{i} \right ), \textrm{where } k_{i}\in \left \{ 1,2,\cdots ,K_{i} \right \},
	\end{equation}
	and then feed them into the mixing network, producing the value of $Q_{tot}^{s}$. This process can be denoted by:
	\begin{equation}
	Q_{tot}^{s} = f(\textbf{s},\widehat{Q}_{1},\cdots,\widehat{Q}_{N};\omega_{mix})
	\end{equation}
	We update all agents' continuous policies $\mu_{i}$ $(i\in N)$ by maximizing $Q_{tot}^{s}$ with parameters $\omega_{i}$ and $\omega_{mix}$ fixed, the gradient can be written as:
	\begin{equation}
	\begin{split}
	&\nabla_{\theta _{i}}{l}(\theta _{i})=\\ &\mathbb{E}_{s,\vec{k} \sim \mathcal{D}}[\nabla_{\theta _{i}}\mu _{k_{i}}( o_{i})\nabla_{x_{k_{i}}}Q_{tot}^{s}(\textbf{s}, \vec{k}, \vec{x}_{k}; \omega )\mid _{x_{k_{i}}=\mu_{k_{i}}(o_{i})}]
	\end{split}
	\end{equation}
	In this way we can update the policies of continuous parameters for all the different agents and different discrete actions in one single training step. 
	
	However, this algorithm may result in high computational complexity during both training and execution phases: every time we compute the joint action value $Q_{tot}$, we need to compute continuous parameters for all the K discrete actions of all the agents. This is particularly severe when the discrete part of hybrid action space has a large dimension. The same problem exists in original P-DQN as well. 
	\subsection{Deep Multi-Agent Hierarchical Hybrid Q-Networks (Deep MAHHQN)}
	
	To address the problem we mentioned in previous section, we propose another novel algorithm Deep MAHHQN, as inspired by Hierarchical Learning~\cite{DBLP:conf/nips/KulkarniNST16,Tang2018HierarchicalDM}.
	
\begin{figure}[t]
		\centering
		\includegraphics[width=0.99\linewidth]{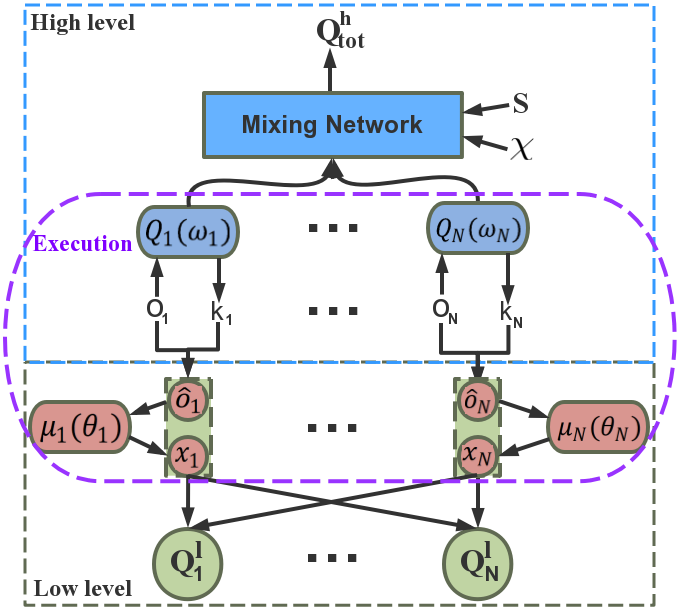}
		\caption{The overall Deep MAHHQN structure}
		\label{fig:98}
	\end{figure}

	The overall structure of Deep MAHHQN is illustrated in Figure \ref{fig:98}. Deep MAHHQN consists of a high-level network coordinating the learning over joint discrete actions and a low-level network for learning the coordinated policy over the continuous parameters. We train the high-level network and low-level network separately and both of them follow the centralized training but decentralized execution paradigm.
	
	Different from Deep MAPQN, when choosing hybrid actions, a deep MAHHQN agent first chooses a discrete action through high-level network and then decides the corresponding continuous parameters conditioning on the given discrete action and individual observation through low-level network. This is consistent with human's decision-making process since in real world humans usually tend to decide what to do before deciding to what extent to do it.
	
	As for the high-level network, each agent utilizes the basic DQN~\cite{Mnih2015HumanlevelCT} framework to select a discrete action $k_{i}$ given its own observation $o_{i}$. The high-level policies among agents over discrete actions are coordinated by using the mixing network structure~\cite{Rashid2018QMIXMV}. Importantly, unlike Qmix which only takes global state $\textbf{s}$ as inputs for the hypernetworks to produce weights of mixing networks, we further consider low-level network's current policies for each agent. This information is critical for coordinating agents' discrete actions since the hybrid action spaces are highly correlated in determining the global optimal policies. Concretely, at each training step, we first calculate the continuous parameters $x_{i}$ related to each agent's high level action $k_{i}$ using the current low-level network's policy (i.e., $x_{i}=\mu_{i}(\varphi (o_{i}, k_{i}))$ ). Then we combine them (i.e., $\chi = \left \{ x_{1},\cdots,x_{N}\right \}$ ) with the global state $\textbf{s}$ and feed them into the hypernetworks to generate weights for the mixing network. The high-level networks' parameters are updated by minimizing the following loss:
	\begin{equation}
	\mathcal{L}(\omega^{h} ) = \mathbb{E}_{s,\vec{k},r,s'\sim \mathcal{D}}(y_{tot}^{h} - Q_{tot}^{h}(\textbf{s}, \vec{k}, \chi))^{2},
	\end{equation}
	where $\vec{k}$ denotes the joint discrete action sampled from the replay buffer, $y^{tot}=r + \gamma \max_{\vec{k}'}Q_{tot}^{h} (\textbf{s}',\vec{k}', \chi')$, $\chi'$ is a set of continuous parameters from low-level target policies.
	
	For the low-level part, each agent $i$ chooses continuous parameters $x_{i}$ according to its new observation: 
	\begin{equation}
	\widehat{o}_{i} = \varphi (o_{i}, k_{i}),
	\end{equation}
	where $k_{i}$ is the discrete action obtained from the high-level part. In our experiments, we simply concatenate $o_{i}$ and $k_{i}$ as the new observation $\widehat{o}_{i}$. In order to learn a coordinated policy over the corresponding continuous parameters, we apply Multi-agent Actor-critic framework~\cite{Lowe2017MultiAgentAF} and further modify it with a centralized Q function for each agent. Considering $N$ agents with low-level policies parameterized by $\theta =\left \{ \theta _{1},\cdots ,\theta _{N} \right \}$, and let $\mu =\left \{ \mu _{1},\cdots ,\mu _{N} \right \}$ be the set of all agents' low-level policies. The gradient with expected return for agent $i$ with low-level policy $\mu_{i}$ can be written as:
	\begin{equation}
	\begin{split}
	\label{E1}
	&\nabla_{\theta _{i}}{l}(\theta _{i})=\\ &\mathbb{E}_{s,k,x\sim \mathcal{D}}[\nabla_{\theta _{i}}\mu _{i}( \widehat{o_{i}})\nabla_{x_{i}}Q_{i}^{l}(\textbf{s}, k_{1}, x_{1},\cdots ,k_{N}, x_{N})|_{x_{i}=\mu_{i}(\widehat{o_{i}})}]
	\end{split}
	\end{equation}
	Here $Q_{i}^{l}(\textbf{s}, k_{1}, x_{1},\cdots ,k_{N}, x_{N})$ is a centralized action-value function that takes as input the hybrid actions of all agents $(k_{1}, x_{1}),\cdots ,(k_{N}, x_{N})$, in addition to the global state $\textbf{s}$, and outputs the low-level Q-value for agent $i$. The experience replay buffer $\mathcal{D}$ contains the tuples $(\textbf{s},k_{1}, x_{1},\cdots ,k_{N}, x_{N},r_{1},\cdots ,r_{N},\textbf{s}')$.
	The centralized Q function is updated as:
	\begin{equation}
	\begin{split}
	\label{E2}
	\mathcal{L}(\omega_{i}^{l} ) &= \mathbb{E}_{s,k,x,r,s'\sim \mathcal{D}}[y_{i} - Q_{i}^{l}(\textbf{s}, k_{1}, x_{1},\cdots,k_{N}, x_{N})]^{2},\\ &y_{i} = r_{i} + \gamma Q_{i}^{l'} (\textbf{s}', k_{1}', x_{1}',\cdots ,k_{N}', x_{N}')|_{x_{j}'=\mu_{j}'(\widehat{o_{i}}')},
	\end{split}
	\end{equation}
	where $Q_{i}^{l'}$ denotes the low-level target Q network for agent $i$. Note that $k_{i}'$ are derived from high-level target policies. Combining (14) and (15) yields our proposed low-level network.
	
	Compared with Deep MAPQN, Deep MAHHQN only needs to calculate one discrete action with the optimal continuous parameters for each agent. This would significantly reduce the algorithm's computational complexities, which will be validated in our experimental results. Moreover, both low-level and high-level training frameworks of Deep MAHHQN are augmented with extra information about policies of other agents and policies of other action levels. In hybrid action environments, we expect that such kind of communication would better alleviate the non-stationary issue of the environments.
	
	When training Deep MAHHQN, we let the low-level network train alone for $m$ steps and then start training high-level and low-level network together. The main reason is that the associated low-level continuous parameters play an important role when our high-level network computes the value of $Q_{tot}^{h}$. Otherwise, the gradients for high-level network can be very noisy and misleading since the low-level policies are still exploring at a high rate at the very beginning.
	
	\section{Experimental Results}
	In this section, we evaluate our algorithms in 1) the standard benchmark game HFO, 2) 3v3 mode in a large-scale online video game $Ghost$ $Story$. We compare our algorithms, Deep MAPQN and Deep MAHHQN, with independent P-DQN in all our experiments.
	\subsection{Experiments with Half Field Offense (HFO)}
	\subsubsection{Environment Settings}
	Half field Offense (HFO) is an abstraction of full RoboCup 2D game. The environment features in a hybrid action space. Previous work~\cite{Hausknecht2016DeepRL,Wang2018ExponentiallyWI,DBLP:conf/aaaiss/WeiWL18} applied RL to the single-agent version of HFO, letting one agent try to goal without a goalkeeper (1v0). In this paper, we apply our proposed algorithms on the challenging multi-agent problems of HFO , which includes \textbf{1v2} (two agents with the shared objective of defending the goal against one opponent) and \textbf{2v1} (two agents with the shared objective of scoring the goal against one goalie). The opponent (or goalie) we play against are all built-in handcoded agents.
	
	We use the high level feature set and each agent's observation is a 21-d vector consisting of its position and orientation; distance and angle to the ball and goal; an indicator if the agent can kick etc. A full list of state information can be found at the official website \url{https://github.com/mhauskn/ HFO/blob/master/doc/manual.pdf}.
	
	In our experiments, we use a discrete-continuous hybrid action space. The full set of hybrid actions is:
	$\textbf{Kick To } (target_{x}, target_{y},speed)\textbf{; Move To }(target_{x},$
	$target_{y})\textbf{; Dribble To } (target_{x}, target_{y})\textbf{; Intercept()}$. 
	Valid values for $target_{x,y}\in \left [ -1,1 \right ]$ and $speed\in [0,3]$. In this settings, acting randomly is almost unlikely to score or successfully defend the goal and the exploration task proves too difficult to gain traction on a reward that only consists of scoring goals. Thus, we do reward engineering to our two tasks respectively to alleviate the sparse reward problem, which will be described in details in following sections.
	\begin{figure}[b]
		\centering
		\includegraphics[width=0.99\linewidth]{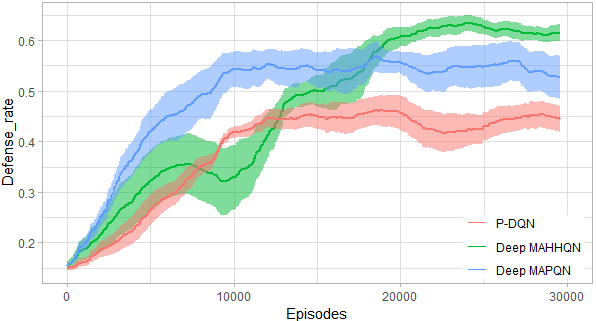}
		\caption{Successfully defense rates for Deep MAPQN, Deep MAHHQN and P-DQN in 1v2 defense mode of HFO}
		\label{fig:112}
	\end{figure}
	
	\subsubsection{1v2 Defense} 
	In the defending scenario, our models control two agents with the shared objective of defending the goal. Note that for defensive agents, only two actions \textbf{Move To} and \textbf{Intercept} are applicable since the defensive players do not control the ball. The reward for each time step is calculated as a weighted sum of the following three types of statistics:
	\begin{itemize}
		\item Move to ball reward: A reward proportional to the change in distance between the agent and the ball.
		\item Punishment for no agent in goal area. We add a punishment if there's no defensive agent in the goal area.
		\item Bonus points for game result. Agents will get extra positive points if they successfully defend the goal and vice versa.
	\end{itemize}
	
	We can see from Figure \ref{fig:112} that both Deep MAPQN and Deep MAHHQN achieve much higher performance than P-DQN. This demonstrates the benefits of explicitly coordinating the joint hybrid policies among agents. In addition, Deep MAHHQN is found to outperform Deep MAPQN after convergence. We attribute this to the improved communication between different agents when we do centralized training for Deep MAHHQN. We further examine the learned behaviours of policies in order to better understand the methods. The agents of Deep MAPQN and Deep MAHHQN tend to play different roles automatically when defending the ball: one agent moves directly to the goal area and act like a goalkeeper while the other one approaches the offensive player trying to capture the ball. In contrast, the agents of P-DQN tend to approach the offensive player or move to the goal area together without cooperating with an appropriate division of their roles. A video of our learned policies may be viewed at \url{https://youtu.be/ndJYZFL5BxE}.

	\subsubsection{2v1 Offense}
	We further evaluate our algorithms on 2v1 offense mode which have larger action space and the coordination task is expected to be more challenging. In this scenario, our models control two agents aiming to score a goal against one goalie. We add a discrete action \textbf{Shoot()} to our action list and there are totally five types of actions for this experiment. The reward for each time step is calculated as a weighted sum of the following types of statistics:
	\begin{itemize}
		\item Move to ball reward: Similar to last section.
		\item Kick to goal reward: A reward proportional to change in distance between the ball and the center of the goal.
		\item Additional punishment: To avoid long shot, we add an additional punishment if the ball is too far away from both agents when it is still outside the goal area.
		\item Bonus points for game result: Agents will get extra points if they successfully score a goal.
	\end{itemize}
	\begin{figure}[b]
		\centering
		\includegraphics[width=0.99\linewidth]{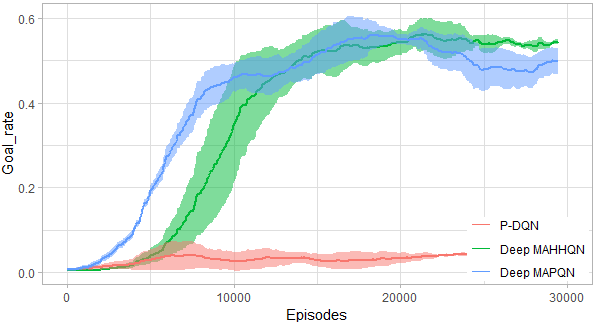}
		\caption{Goal rates for Deep MAPQN, Deep MAHHQN and P-DQN in 2v1 offense mode of HFO}
		\label{fig:110}
	\end{figure}
	As shown in Figure \ref{fig:110}, in this challenging offense problem, independent P-DQN fails to learn the coordination policy. In practice, we observe that independent P-DQN agents first try to approach the ball, then dribble it to somewhere and stop without shooting. A primary reason for this may be the lack of information of other teammates. In comparison, our proposed methods Deep MAPQN and Deep MAHHQN perform much better. Similar to the previous defense setting, the Deep MAPQN agents learn policies a little faster than Deep MAHHQN in terms of the number of time steps required. The underlying reason is that in Deep MAPQN we update policies of continuous parameters associated with all the discrete actions at one single training step. However, such a framework also introduces huge computational complexity since each training step of Deep MAPQN occupies much more computing resources than Deep MAHHQN. This problem can be seen more clearly in our second game (Ghost Story) with more agents.
	
	To summarize, both Deep MAPQN and Deep MAHHQN outperform independent P-DQN, and Deep MAHHQN seems to perform slightly better than Deep MAPQN. In our next scenario, we test our algorithms in a complicated practical environment with larger hybrid action spaces and state spaces.
	
	\subsection{Experiments with Ghost Story}
	We evaluate the proposed algorithms with \textit{Ghost Story} (or \textit{QianNvYouHun}) -- a fantasy massive multiplayer online role-playing game (MMORPG) from \textit{NetEase}, in which each of the learning agents controls an individual hero. We performed our evaluation in the "3v3" game mode, where 3 hero agents cooperate with each other to fight against the other 3 built-in AI on the opposite side. At each time step, every hero can choose to move or use one of its own skills. The game ends when all 3 heroes on the same side are killed. 
	\begin{figure}[t]
		\centering
		\includegraphics[width=0.99\linewidth]{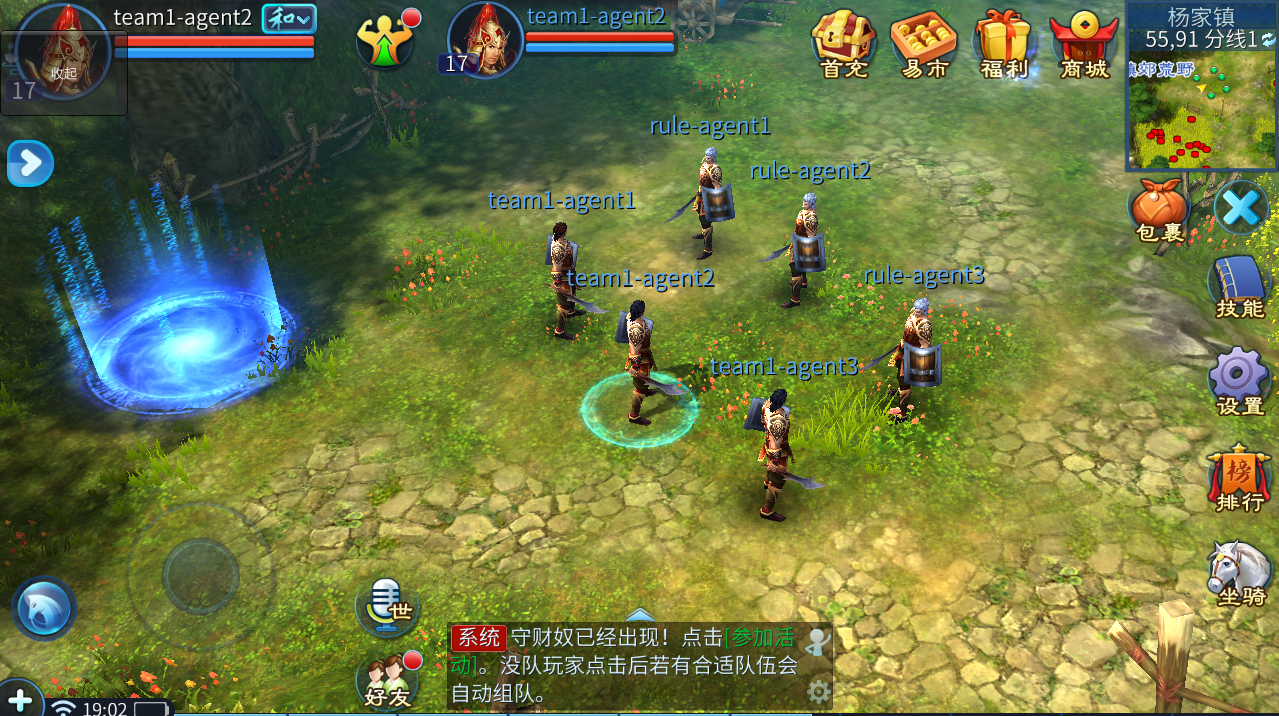}
		\caption[Figure 5]{A screen shot of a 3v3 game mode in $Ghost$ $Story$}
		\label{fig:qnyh}
	\end{figure}
	
	The observation for each agent is a 97-d feature vector which is manually constructed using output from the game engine. Concretely, these features consist of some basic properties of the agents: agent's Health Point ($\textsl{HP}$), value range of attack, value range of defense\footnote{The agent's true attack or defense value is uniformly distributed on the given value range.}, carried skills and Cool Down ($\textsl{CD}$), carried buffs\footnote{Buffs can be beneficial, such as increase agent's own defense value, or harmful, such as decrease its own attack value, but buffs can only exist for a short time period.}, relative positions etc. 
	
	We simplify the actions of each hero into five hybrid action types: $Move(x,y)$, move to a relative position(x,y); and four skills, $Tanlang(x,y)$, $Siguai(x,y)$, $Tianshou()$ and $Hegu(x,y)$. When a hero player chooses to use a skill, the enemies near the relative position $(x,y)$ will be attacked or added some buffs depending on the specific types of selected skills. Skill $Tianshou$ has no parameters since it functions as adding a buff on the hero agent itself. More details of our experimental settings can be found at \url{https://bit.ly/2Eaci2X}.
	
	At each time step, each agent receives a joint reward consists of four parts: (1) the change in $\textsl{HP}$ for all heroes; (2) a punishment for the agents which did not do anything (e.g. No enemy is near the skill's target point; the selected skill's $\textsl{CD}$ is not zero ); (3) small bonus points for killing an enemy hero and huge bonus points for winning the game.
	\begin{figure}[t]
		\centering
		\includegraphics[width=0.99\linewidth]{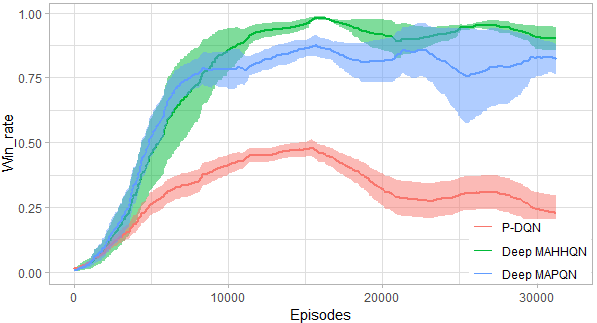}
		\caption[Figure 6]{Win rates for Deep MAPQN, Deep MAHHQN and P-DQN in 3v3 mode of $Ghost$ $Story$}
		\label{fig:111}
	\end{figure}
	
	Figure \ref{fig:111} shows our experimental results. Clearly we can see independent P-DQN fails to learn a coordinated policy that can consistently defeat the enemies. Both Deep MAPQN and Deep MAHHQN outperform independent P-DQN learning method and get a win rate over 75$\%$. Moreover, we find Deep MAHHQN method performs slightly better than Deep MAPQN, which further validates that the improved communication between agents using information about policies of other action levels can better stabilize the centralized training process. It indicates that Deep MAHHQN framework provides a better way to handle problems with large hybrid action space and number of agents. A final note is that the actual training time of Deep MAPQN is about three days while Deep MAHHQN takes less than one day to train on the same NVidia Geforce GTX 1080Ti GPU. This is quite reasonable since each MAPQN agent need to compute continuous parameters for all the discrete actions at each single training step while in Deep MAHHQN we only need to calculate low level continuous parameters associated with the selected optimal high-level discrete action.

	\section{Conclusion}
	This paper should be seen as the first attempt at applying deep reinforcement learning in cooperative multi-agent settings with discrete-continuous hybrid action spaces. Two novel approaches are proposed under the paradigm of centralized training and decentralized execution. The experimental results show their superiority to independent parameterized Q-learning method under both the standard testbed HFO and a large-scale MMORPG game. As future work, we wish to extend our algorithms to competitive multi-agent settings, and conduct additional experiments with more agents and larger hybrid action space to further investigate the difference of performance between our two proposed algorithms.

	\bibliographystyle{named}
	\bibliography{ijcai19}

\begin{thebibliography}{}

\bibitem[\protect\citeauthoryear{Cao \bgroup \em et al.\egroup
  }{2013}]{Cao2013AnOO}
Yongcan Cao, Wenwu Yu, Wei Ren, and Guanrong Chen.
\newblock An overview of recent progress in the study of distributed
  multi-agent coordination.
\newblock {\em IEEE Transactions on Industrial Informatics}, 9:427--438, 2013.

\bibitem[\protect\citeauthoryear{Foerster \bgroup \em et al.\egroup
  }{2016}]{DBLP:conf/nips/FoersterAFW16}
Jakob~N. Foerster, Yannis~M. Assael, Nando de~Freitas, and Shimon Whiteson.
\newblock Learning to communicate with deep multi-agent reinforcement learning.
\newblock In {\em Proceedings of NeurIPS}, pages 2137--2145, 2016.

\bibitem[\protect\citeauthoryear{Hausknecht and
  Stone}{2016}]{Hausknecht2016DeepRL}
Matthew~J. Hausknecht and Peter Stone.
\newblock Deep reinforcement learning in parameterized action space.
\newblock {\em In Proceedings of ICLR}, pages 861--868, 2016.

\bibitem[\protect\citeauthoryear{Hausknecht}{2016}]{Hausknecht2016HalfFO}
Matthew~J. Hausknecht.
\newblock Half field offense : An environment for multiagent learning and ad
  hoc teamwork.
\newblock In {\em Proceedings of ALA}, pages 1391--1398, 2016.

\bibitem[\protect\citeauthoryear{Kulkarni \bgroup \em et al.\egroup
  }{2016}]{DBLP:conf/nips/KulkarniNST16}
Tejas~D. Kulkarni, Karthik Narasimhan, Ardavan Saeedi, and Josh Tenenbaum.
\newblock Hierarchical deep reinforcement learning: Integrating temporal
  abstraction and intrinsic motivation.
\newblock In {\em Proceedings of NeurIPS}, pages 3675--3683, 2016.

\bibitem[\protect\citeauthoryear{Lillicrap \bgroup \em et al.\egroup
  }{2016}]{Lillicrap2016ContinuousCW}
Timothy~P. Lillicrap, Jonathan~J. Hunt, Alexander Pritzel, Nicolas Heess, Tom
  Erez, Yuval Tassa, David Silver, and Daan Wierstra.
\newblock Continuous control with deep reinforcement learning.
\newblock {\em Proceedings of ICLR}, pages 1052--1059, 2016.

\bibitem[\protect\citeauthoryear{Lowe \bgroup \em et al.\egroup
  }{2017}]{Lowe2017MultiAgentAF}
Ryan Lowe, Yi~Wu, Aviv Tamar, Jean Harb, Pieter Abbeel, and Igor Mordatch.
\newblock Multi-agent actor-critic for mixed cooperative-competitive
  environments.
\newblock In {\em Proceedings of NeurIPS}, pages 6382--6393, 2017.

\bibitem[\protect\citeauthoryear{Masson \bgroup \em et al.\egroup
  }{2016}]{DBLP:conf/aaai/MassonRK16}
Warwick Masson, Pravesh Ranchod, and George Konidaris.
\newblock Reinforcement learning with parameterized actions.
\newblock In {\em Proceedings of AAAI}, pages 1934--1940, 2016.

\bibitem[\protect\citeauthoryear{Mnih \bgroup \em et al.\egroup
  }{2013}]{Mnih2013PlayingAW}
Volodymyr Mnih, Koray Kavukcuoglu, David Silver, Alex Graves, Ioannis
  Antonoglou, Daan Wierstra, and Martin~A. Riedmiller.
\newblock Playing atari with deep reinforcement learning.
\newblock {\em CoRR}, abs/1312.5602, 2013.

\bibitem[\protect\citeauthoryear{Mnih \bgroup \em et al.\egroup
  }{2015}]{Mnih2015HumanlevelCT}
Volodymyr Mnih, Koray Kavukcuoglu, David Silver, Andrei~A. Rusu, Joel Veness,
  Marc~G. Bellemare, Alex Graves, Martin~A. Riedmiller, Andreas Fidjeland,
  Georg Ostrovski, Stig Petersen, Charles Beattie, Amir Sadik, Ioannis
  Antonoglou, Helen King, Dharshan Kumaran, Daan Wierstra, Shane Legg, and
  Demis Hassabis.
\newblock Human-level control through deep reinforcement learning.
\newblock {\em Nature}, 518:529--533, 2015.

\bibitem[\protect\citeauthoryear{Peng \bgroup \em et al.\egroup
  }{2017}]{DBLP:journals/corr/PengYWYTLW17}
Peng Peng, Quan Yuan, Ying Wen, Yaodong Yang, Zhenkun Tang, Haitao Long, and
  Jun Wang.
\newblock Multiagent bidirectionally-coordinated nets for learning to play
  starcraft combat games.
\newblock {\em CoRR}, abs/1703.10069, 2017.

\bibitem[\protect\citeauthoryear{Rashid \bgroup \em et al.\egroup
  }{2018}]{Rashid2018QMIXMV}
Tabish Rashid, Mikayel Samvelyan, Christian~Schr{\"o}der de~Witt, Gregory
  Farquhar, Jakob~N. Foerster, and Shimon Whiteson.
\newblock Qmix: Monotonic value function factorisation for deep multi-agent
  reinforcement learning.
\newblock In {\em Proceedings of ICML}, pages 4292--4301, 2018.

\bibitem[\protect\citeauthoryear{Tang \bgroup \em et al.\egroup
  }{2018}]{Tang2018HierarchicalDM}
Hongyao Tang, Jianye Hao, Tangjie Lv, Yingfeng Chen, Zongzhang Zhang, Hangtian
  Jia, Chunxu Ren, Yan Zheng, Changjie Fan, and Li~Wang.
\newblock Hierarchical deep multiagent reinforcement learning.
\newblock {\em CoRR}, abs/1809.09332, 2018.

\bibitem[\protect\citeauthoryear{Wang \bgroup \em et al.\egroup
  }{2018}]{Wang2018ExponentiallyWI}
Qing Wang, Jiechao Xiong, Lei Han, Peng Sun, Han Liu, and Tong Zhang.
\newblock Exponentially weighted imitation learning for batched historical
  data.
\newblock In {\em Proceedings of NeurIPS}, pages 6291--6300, 2018.

\bibitem[\protect\citeauthoryear{Wei \bgroup \em et al.\egroup
  }{2018a}]{DBLP:conf/aaaiss/WeiWFL18}
Ermo Wei, Drew Wicke, David Freelan, and Sean Luke.
\newblock Multiagent soft q-learning.
\newblock In {\em Proceedings of AAAI}, pages 1729--1736, 2018.

\bibitem[\protect\citeauthoryear{Wei \bgroup \em et al.\egroup
  }{2018b}]{DBLP:conf/aaaiss/WeiWL18}
Ermo Wei, Drew Wicke, and Sean Luke.
\newblock Hierarchical approaches for reinforcement learning in parameterized
  action space.
\newblock In {\em Proceedings of AAAI}, pages 3211--3218, 2018.

\bibitem[\protect\citeauthoryear{Xiong \bgroup \em et al.\egroup
  }{2018}]{Xiong2018ParametrizedDQ}
Jiechao Xiong, Qing Wang, Zhuoran Yang, Peter~P Sun, Lei Han, Yang Zheng, Haobo
  Fu, Tong Zhang, Ji~Liu, and Hao Liu.
\newblock Parametrized deep q-networks learning: Reinforcement learning with
  discrete-continuous hybrid action space.
\newblock {\em CoRR}, abs/1810.06394, 2018.

\bibitem[\protect\citeauthoryear{Ye \bgroup \em et al.\egroup
  }{2015}]{DBLP:journals/sensors/YeZY15}
Dayong Ye, Minjie Zhang, and Yun Yang.
\newblock A multi-agent framework for packet routing in wireless sensor
  networks.
\newblock {\em Sensors}, 15(5):10026--10047, 2015.

\end{thebibliography}
	
\end{document}